\documentclass[pdflatex,sn-mathphys-num]{sn-jnl}

\usepackage{graphicx}%
\usepackage{multirow}%
\usepackage{amsmath,amssymb,amsfonts}%
\usepackage{amsthm}%
\usepackage{mathrsfs}%
\usepackage[title]{appendix}%
\usepackage{xcolor}%
\usepackage{textcomp}%
\usepackage{manyfoot}%
\usepackage{booktabs}%
\usepackage{algorithm}%
\usepackage{algorithmicx}%
\usepackage{algpseudocode}%
\usepackage{listings}%
\usepackage{svg}
\usepackage{pdfpages}

\theoremstyle{thmstyleone}%
\theoremstyle{thmstyletwo}%

\theoremstyle{thmstylethree}%

\begin{document}
\title[Article Title]{An Open-Source Monitoring Framework for Data Exploration and Progress Tracking in Multi-Center Radiology Studies}


\author*[1,2]{\fnm{Markus} \sur{Bujotzek}}\email{markus.bujotzek@dkfz-heidelberg.de}
\author[1]{\fnm{Jonas} \sur{Scherer}}
\author[1,3]{\fnm{Stefan} \sur{Denner}}
\author[1]{\fnm{Peter} \sur{Neher}}
\author[1,2]{\fnm{Benjamin} \sur{Hamm}}
\author[1]{\fnm{Lorenz} \sur{Feineis}}
\author[1]{\fnm{Ünal} \sur{Akünal}}
\author[4]{\fnm{Andreas} \sur{Bucher}}
\author[5,6]{\fnm{Tobias} \sur{Penzkofer}}
\author[1,2,3]{\fnm{Klaus} \sur{Maier-Hein}}

\affil[1]{\orgdiv{Division of Medical Image Computing}, \orgname{Germany Cancer Research Center}, \city{Heidelberg}, \postcode{69120}, \country{Germany}}

\affil[2]{\orgdiv{Medical Faculty}, \orgname{University of Heidelberg}, \city{Heidelberg}, \postcode{69120}, \country{Germany}}


\affil[3]{\orgdiv{Faculty for Computer Science}, \orgname{University of Heidelberg}, \city{Heidelberg}, \postcode{69120}, \country{Germany}}

\affil[4]{\orgdiv{Institute for Diagnostic and Interventional Radiology}, \orgname{University Hospital Frankfurt}, \city{Frankfurt}, \postcode{60590}, \country{Germany}}

\affil[5]{\orgdiv{Department of Radiology}, \orgname{Charite Universitätsmedizin Berlin}, \city{Berlin}, \postcode{10117}, \country{Germany}}

\affil[6]{\orgdiv{Berlin Institute of Health}, \city{Berlin}, \postcode{10178}, \country{Germany}}


\abstract{
\textbf{Purpose.}
Multi-center studies are crucial for advancing medical and radiological research. Data exploration, collaboration discovery, and study progress monitoring are essential for maximizing their potential. However, in practice these processes often rely on manual communication and shared tables, which quickly become outdated and hinder efficient coordination in large distributed studies. This highlights the need for dedicated monitoring solutions that provide transparent and up-to-date insights into study progress.

\textbf{Methods.}
We propose a lightweight, open-source monitoring architecture for multi-center studies based on the widely used Grafana–Prometheus stack. The framework collects aggregated monitoring metrics from distributed study sites and visualizes them through configurable dashboards. As a real-world deployment example, the framework is integrated into the medical imaging platform Kaapana and evaluated within a large multi-center research network.

\textbf{Results \& Conclusion.}
By deploying our solution within the Germany-wide RACOON consortium, we demonstrate its ability to enable privacy-preserving data exploration and study progress monitoring across all 38 German university clinics. The monitoring framework supports transparent coordination of distributed research activities and can facilitate more efficient management of large-scale multi-center studies. The source code and Kaapana integration are publicly available at \hyperlink{https://github.com/MIC-DKFZ/study-monitoring-kaapana}{https://github.com/MIC-DKFZ/study-monitoring-kaapana}.
}




\maketitle

\section{Background}\label{sec1}
Multi-center studies play a crucial role in advancing radiological and general medical research.
Leveraging medical expertise and data from multiple institutions enables larger, more diverse and therefore more representative patient cohorts improving generalizability and applicability of clinical findings \cite{dashevsky2018multicenter}.
These benefits of multi-centric collaborations have been proven successfully in several studies in the domain of radiology \cite{bucher2025prognostic} as well as AI-assisted radiology \cite{pati2022federated,bujotzek2025real}.

Despite the apparent advantages of multi-center studies\cite{dashevsky2018multicenter}, they come with significant hurdles. A key issue is data exploration: while clinics may join large-scale consortia, identifying available data and selecting suitable collaborators remains complex. This information is often buried in private emails, making it inaccessible to new collaborators and prone to becoming outdated.
In addition, even after collaborations are established, tracking study progress, such as acquired and analyzed patient enrollment, workflow execution, and milestone achievements, is tedious. Current methods rely on private emails or manually updated shared tables, both error-prone as well as easily outdated due to manual data entry risks, and simply infeasible for large-scale studies involving hundreds of medical research centers.
These hurdles underscore the urgent need for dedicated tools facilitating transparent data exploration and up-to-date study monitoring in multi-centric research.

Existing tools in the healthcare domain address data storage \cite{pozamantir2010web, marcus2007extensible}, management \cite{hayashi2024brainlife, das2012loris}, sharing \cite{poline2023data, marcus2007extensible} and documentation \cite{kessel2012web} in multi-center studies.
While effective within their specific scopes, these solutions are often developed as isolated, project- or data-source-specific implementations \cite{pozamantir2010web, engels2024sample}, limiting their generalizability and integrability with existing healthcare IT infrastructure, or rely on data sharing compromising privacy \cite{poline2023data}. Instead of leveraging and adopting well-established open-source solutions, they frequently reinvent the wheel, introducing entirely new platforms that require significant customization \cite{pozamantir2010web, hayashi2024brainlife, das2012loris, poline2023data, kessel2012web, marcus2007extensible, maier2023profile}.
While the adoption of open-source monitoring technology stacks in healthcare remains sparse \cite{ccalhan2021ehealth}, other domains leverage these tools to track various systems \cite{betke2017real, jani2024unified, mehdi2023unleashing, al2021realtime}.
A widely established and applied monitoring stack consists of Grafana \cite{chakraborty2021grafana} for visualizing metrics and Prometheus \cite{turnbull2018monitoring} as a data source provider.
These open-source monitoring solutions are extensively used across various domains, including high-performance computing \cite{betke2017real}, microservice architectures \cite{jani2024unified}, servers, databases, IoT systems, and business applications \cite{mehdi2023unleashing}, up to to niche applications like monitoring environmental conditions in greenhouses \cite{al2021realtime}. Despite their widespread adoption, their open-source implementation and application into distributed and multi-center healthcare IT infrastructure remains largely unexplored.

\begin{figure}[t]
    \centering
    \includegraphics[width=\textwidth]{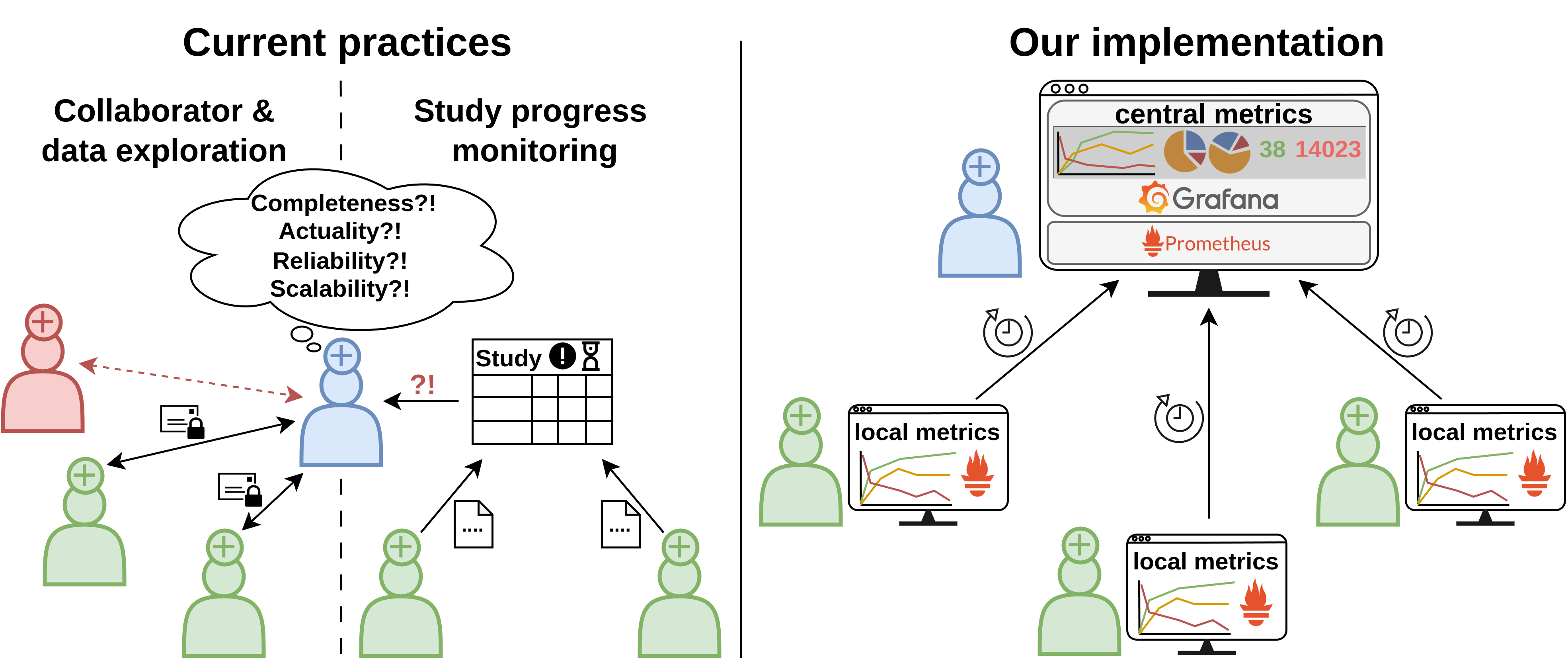}
    \caption{\textbf{Problems in current practices and our implementation as a solution for data exploration and progress monitoring of multi-center studies}. Current practices often rely on private emails for collaborator and data exploration, and on shared tables for study monitoring; both excluding potential new collaborators, prone to errors, quickly outdated, and not scalable. To overcome these hurdles, we propose our lightweight, configurable, web-based monitoring framework built on the open-source monitoring stack Grafana and Prometheus.} 
    \label{fig:problem_statement}
\end{figure}

Building upon this successful and widespread adoption \cite{betke2017real, jani2024unified, mehdi2023unleashing, al2021realtime}, we propose a lightweight, open-source, web-based solution for data exploration and study progress monitoring in multi-center studies.
Our implementation leverages a well-established, web-based monitoring technology stack, utilizing Prometheus for data provisioning and Grafana dashboards for intuitive visualization, (figure \ref{fig:problem_statement}), and adopts these tools for efficient integration and monitoring of radiological studies.
To demonstrate the integrability of our proposed implementation, we further integrate it seamlessly into the medical image processing platform Kaapana\footnote{\url{https://www.kaapana.ai/}} \cite{akunal2025kaapana}, enabling multi-center data exploration and study monitoring in radiology.
Finally, we validate the functionality of our implementation and integration into Kaapana by deploying it within the RAdiological COOperative Network (RACOON\footnote{\url{https://racoon.network/}}) consortium. This deployment allows clinical researchers to explore collaboration partners and their data in a privacy-preserving manner, while monitoring study progress through image-related metrics and workflow execution statistics.
Although the deployed and evaluated implementation in this work is integrated into the Kaapana platform, the proposed monitoring architecture is platform-independent and can be applied to any research infrastructure exposing monitoring metrics through standard interfaces.

In summary, our key contributions are:
\begin{itemize}
    \item[--] We present a lightweight, open-source monitoring architecture for data exploration and progress tracking in multi-center radiology studies, built on the widely adopted Grafana–Prometheus monitoring stack.
    \item[--] We provide a flexible and configurable framework that enables privacy-preserving monitoring of distributed research infrastructures through aggregated monitoring metrics.
    \item[--] We demonstrate the practical applicability of the framework through integration into the medical image processing platform Kaapana.
    \item[--] We showcase its real-world deployment within the RACOON consortium, enabling data exploration, collaboration partner discovery, and study progress monitoring across all German university clinics.
    \item[--] The source code, along with the Kaapana integration, is publicly available for use and adoption at \hyperlink{https://github.com/MIC-DKFZ/study-monitoring-kaapana}{https://github.com/MIC-DKFZ/study-monitoring-kaapana}.
\end{itemize}

\section{Materials and Methods}\label{sec2}

\subsection{Requirements}\label{sec:2_1_requirements}
To effectively support data and collaborator exploration as well as study progress monitoring in multi-center studies, our implementation must fulfill the following requirements:
\begin{itemize}
    \item[] \textbf{R1 -- Client-server architecture:} Given the distributed nature of multi-center studies, the system must support a distributed infrastructure, with communication between clients deployed at local sites and a central server visualizing aggregated metrics.
    \item[] \textbf{R2 -- Open-source and framework-agnostic:} The implementation needs to use a widely adopted open-source technology stack and a standardized interface for metric requests, facilitating easy integration into diverse healthcare infrastructures, broad usability and a generalizing deployment.
    \item[] \textbf{R3 -- Privacy-preserving metrics:} The system requires to ensure patient's data privacy to enable exploration and monitoring by sharing only aggregated metric values rather than raw data.
    \item[] \textbf{R4 -- Configurable metrics:} Due to the diverse demands of multi-center studies, the system must provide flexibility to configure the collection of metrics and monitored tools at local sites, as well as the visualization of the monitoring metrics at the central site.
    \item[] \textbf{R5 -- Reliable and up-to-date metrics:} The provisioning of data exploration and study monitoring information should not rely on humans to read and communicate certain metrics. Instead, metrics have to be automatically requested from monitored systems and sent regularly to the central server site for up-to-date metrics.
    \item[] \textbf{R6 -- Study separation:} Detailed study progress metrics can reveal sensitive insights or intellectual property (IP) of clinical researchers. To mitigate this risk, the system requires a strict separation of monitoring metrics across different multi-centric studies within one deployment.
\end{itemize}

\subsection{Technical Background}
The presented data exploration and study progress monitoring tool is built upon the well-established, open-source monitoring stack comprising Grafana and Prometheus, and integrated into the Kaapana platform.

Grafana is a widely used, open-source visualization and alerting tool for time-series data. It supports multiple data sources, and allows users to create interactive dashboards for real-time monitoring.
Prometheus is an open-source time-series database and monitoring tool for high-dimensional metric collection. It continuously scrapes, stores, and processes real-time metric data, integrating effortlessly with Grafana as a data provisioning tool.

Kaapana is an open-source medical imaging platform, enabling secure AI deployment, federated processing, and multi-center collaboration while preserving data privacy. Its modular, containerized design ensures scalability, interoperability, and streamlined end-to-end research workflows in radiology and beyond.
Integrating our tool into Kaapana ensures its applicability for multi-center study monitoring in radiology.
The proposed monitoring architecture is independent of Kaapana and relies solely on containerized monitoring tools and standardized metric endpoints compatible with Prometheus. Consequently, any system capable of deploying containers and exposing metrics through such endpoints can adopt the framework with minimal integration effort.

\subsection{Data Exploration and Study Monitoring Framework}
\begin{figure}[t]
    \centering
    \includegraphics[width=\textwidth]{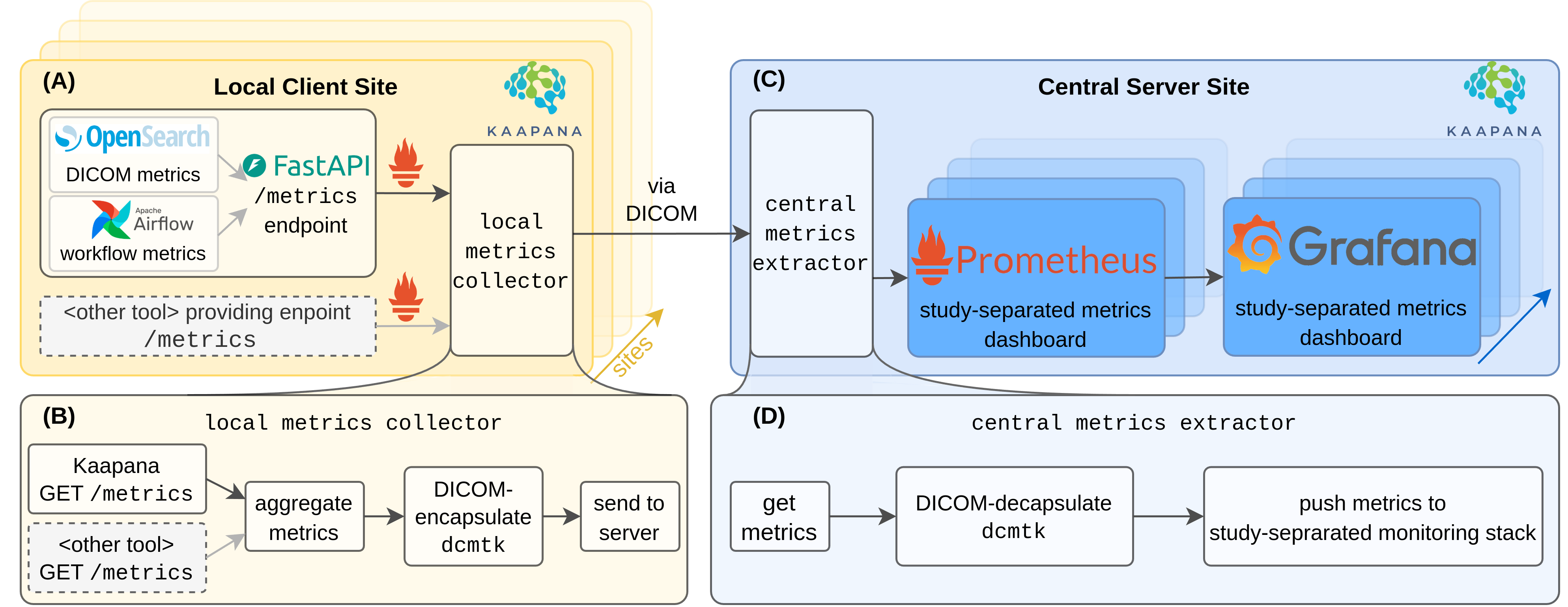}
    \caption{\textbf{Implementation of data exploration and progress monitoring for multi-centric studies within the Kaapana platform.} Monitoring metrics are collected at local client sites (A) via the \texttt{local-metrics-collector} workflow (B) and transmitted to the central server (C). At the central server site, the \texttt{central-metrics-extractor} workflow (D) processes the received metrics, separates study-specific metrics, and pushes them to the dedicated monitoring stack, consisting of Prometheus for data provisioning and Grafana for metric visualization.} 
    \label{fig:high_level_implementation}
\end{figure}

The proposed data exploration and study monitoring framework follows a client-server architecture, aligning with its purpose of providing insights into multi-centric studies. At local client sites, various monitoring metrics are requested from different monitored tools, aggregated, and shared with the central server.
Once arrived at the central server, these metrics are forwarded to a study-separated monitoring stack based on their study context.
This stack consists of Prometheus for data provisioning and Grafana for visualizing the aggregated monitoring metrics, see figure \ref{fig:high_level_implementation}.

\subsubsection{Metrics Collection at Local Sites}\label{sec:2_2_1_metrics_collection}
In our reference implementation, metrics collection at the local client sites is integrated into the Kaapana platform, see figure \ref{fig:high_level_implementation} A and B.
The monitoring tool requests monitoring metrics from any configured local system exposing a \texttt{/metrics} API endpoint.
Following the paradigm of adopting open-source solutions and ensuring compatibility with the technology stack at the central server, monitored systems expose their metrics in the Prometheus text-based exposition format\footnote{\url{https://prometheus.io/docs/instrumenting/exposition_formats/}}.
With Kaapana serving as base infrastructure, the \texttt{local-metrics-collector} workflow retrieves metrics via HTTP GET requests from the \texttt{/metrics} API endpoints of configured, monitored tools.
To monitor Kaapana itself, its FastAPI backend provides access to various metrics, including DICOM metadata from its metadata management tool Opensearch, workflow execution statistics from its workflow engine Apache Airflow, and available disk space on the deployment infrastructure.

The received metrics per monitored tool and study are aggregated and prepared for secure communication between local and central infrastructure instances.
For Kaapana, we leverage its native support for DICOM-based communication, storage, and I/O operations as an existing communication path, avoiding the need to open additional network ports in clinical IT environments, which would introduce potential security risks.
Using the DCMTK toolkit\footnote{\url{https://dicom.offis.de/en/dcmtk/}} we encapsulate aggregated metrics as DICOM files with modality "OT" before transmitting them to the central server for processing and visualization.

For up-to-date monitoring metrics, we eliminate manual data collection and transmission by automating the \texttt{local-metrics-collector} workflow through scheduled cron jobs, ensuring regular and up-to-date metric updates without human intervention.

\subsubsection{Metrics Presentation as Central Site}
The major part of our presented open-source implementation, which remains independent of the underlying Kaapana infrastructure, operates on the central server, see figure \ref{fig:high_level_implementation} C.
The monitoring stack consists of three key components: a Prometheus Pushgateway, Prometheus, and Grafana.
Once metric data reaches the central server instance, it is pushed to the Prometheus Pushgateway for persistence.
Prometheus then retrieves in configured scraping intervals the latest metric data from the Pushgateway, serving as the data provisioning layer for Grafana.
Grafana acts as the interactive visualization component, displaying the collected metrics through various dashboard panels. This allows researchers to explore, inspect, and monitor study progress, while also enabling the dynamic creation of new panels, leveraging complex queries to uncover deeper insights into multi-center studies.
As Kaapana’s architecture is built on containerized microservices deployed via Kubernetes and templated using Helm charts, our implementation packages the monitoring stack components (Prometheus Pushgateway, Prometheus, and Grafana) along service communication resources and further configurations into a standalone Helm chart.

To ensure study-specific monitoring, we deploy a separate monitoring stack for each study, consisting of independent Prometheus Pushgateway, Prometheus, and Grafana instances. The seamless deployment of these study-specific monitoring resources is facilitated by Kaapana’s support for multi-installable Helm resources, which leverage the global study context defined for each study within the Kaapana platform.
To receive, process, and visualize monitoring metrics in a study-separated manner at the central server, a dedicated monitoring stack must be deployed whenever a new study is created.

Once a study-specific monitoring stack is deployed, incoming metrics are received, processed, and visualized. First, the DICOM-encapsulated metrics sent by the sites are ingested through Kaapana’s general DICOM input pipeline and stored in its PACS. Then, the \texttt{central-metrics-extractor} workflow is automatically triggered, checking to automatically process incoming DICOMs of modality "OT".
As detailed in figure \ref{fig:high_level_implementation} D, the \texttt{central-metrics-extractor} workflow retrieves the DICOM-encapsulated metrics from the PACS and extracts the text-based metrics files using the DCMTK toolkit.
Subsequently, the study-separated metrics are pushed to the corresponding study-specific monitoring stack, where they are provisioned through the Prometheus Pushgateway and Prometheus deployments before being visualized on the study-specific Grafana dashboard.
This setup enables fine-grained, up-to-date data exploration, inspection, and progress monitoring of multi-centric studies.

\section{Results}\label{sec3}
Our implementation adopts an established monitoring stack, including Grafana and Prometheus, to enable data exploration and progress monitoring in multi-centric radiology studies.
We transfer proven open-source monitoring technologies to the application field of radiology and integrate them into the medical imaging platform Kaapana as a base infrastructure, making the monitoring framework readily deployable within existing research environments.

\begin{figure}[t]
    \centering
    \includegraphics[width=\textwidth]{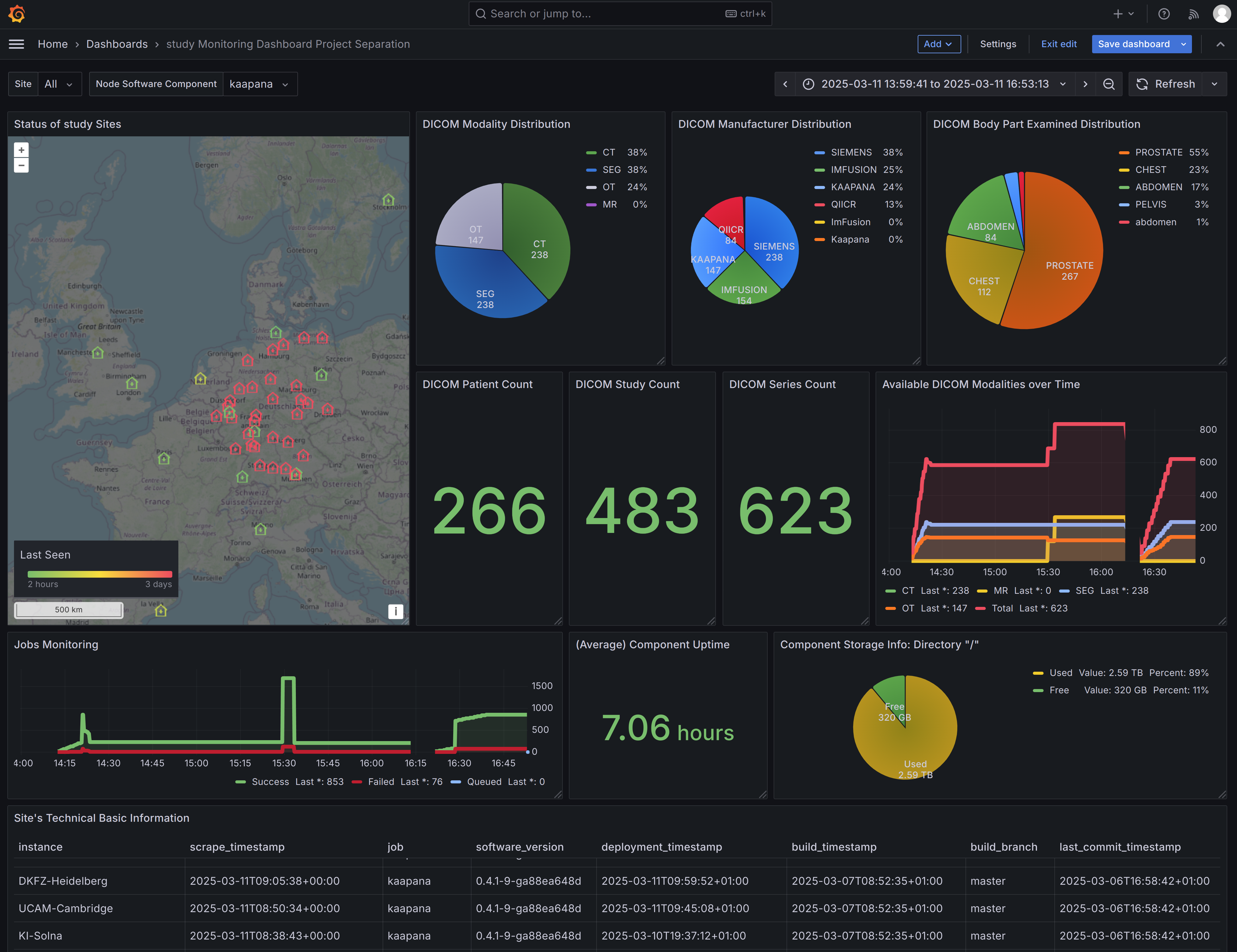}
    \caption{\textbf{Grafana dashboard of the study monitoring tool.}
    The dashboard provides an overview of connected local sites, DICOM data distribution, and system metrics. The left panel visualizes a geographic map of local client sites with status indicators. The center and right panels present key monitoring metrics, including DICOM modality, manufacturer, and body part distributions, as well as patient, study, and series counts. The bottom section offers insights into system metrics, job monitoring, component uptime, storage usage, and installed software versions at local client sites.} 
    \label{fig:screenshot}
\end{figure}

Figure \ref{fig:screenshot} illustrates the Grafana monitoring dashboard of the framework. Beyond visualizing the latest status updates and localization of monitored sites, it enables the visualization of DICOM metadata, including patient, study, and series counts, or modality, manufacturer and examined body part distributions, either aggregated or site-specific.
Additionally, key study progress indicators, such as executed jobs, and technical information, like disk space usage and currently installed software version of local client sites, are monitored.

The default Grafana dashboard we provide offers a flexible starting point, but users of the central server component can configure and visualize aggregated metrics in any preferred panels enabling dynamic configuration possibilities of the central monitoring dashboard.
By default, our tool monitors Kaapana instances at connected local sites as well as the central server instance itself, but any third-party tool at local sites can be integrated and monitored as detailed in section \ref{sec:2_2_1_metrics_collection}. 
Users can filter by local sites and monitored tools across a multi-centric study via drop-down menus \textit{Site} and \textit{Node Software Component}, see figure \ref{fig:screenshot}.

Our tool offers flexible selection of monitored tools, metrics, and visualizations while ensuring study separation to protect researchers' intellectual property. This is achieved through Kaapana’s multi-installable extensions.
A user guide for setting up the monitoring tool in the study-separated way is available in our documentation: \hyperlink{https://github.com/MIC-DKFZ/study-monitoring-kaapana}{https://github.com/MIC-DKFZ/study-monitoring-kaapana}.

A customized version of the Kaapana-based implementation of our tool is actively deployed for monitoring of multi-center studies, most notably within the Germany-wide RACOON consortium.
It facilitates up-to-date data exploration and progress monitoring across \textit{all 38 German university clinics} participating in the network. 
Monitoring metrics are automatically collected at regular intervals from distributed local client sites and aggregated at the central monitoring infrastructure, providing near real-time insights into study progress and infrastructure status.
Beyond monitoring local Kaapana platforms at each site, the monitoring framework extends to additional tools used in RACOON, including Mint Lesion\footnote{https://mint-medical.com/mint-lesion} for medical reporting, as well as ImFusion Labels\footnote{https://www.imfusion.com/products/imfusion-labels} and Fraunhofer Mevis’ CuraMate\footnote{https://www.mevis.fraunhofer.de/en/research-and-technologies/ai-collaboration-toolkit.html}, formerly Satori, for data annotation, by directly querying their exposed \texttt{/metrics} API endpoints.

\subsection{Deployment Experience}
Within the RACOON consortium, the monitoring framework has been applied in a real-world federated learning project involving six university clinics.
In this deployment, infrastructure-related monitoring metrics proved particularly valuable for identifying operational limitations at participating sites. Monitoring available disk space revealed that several resource-limited sites frequently experienced storage shortages, which caused lengthy federated training processes to fail.
The monitoring dashboards enabled rapid identification of such issues and improved coordination between participating institutions during distributed experiments \cite{bujotzek2025real}.

\section{Discussion}\label{sec4}
We present a scalable and open-source solution for data exploration and progress monitoring in multi-centric studies, integrating it into the medical imaging platform Kaapana to make it accessible in radiology and beyond.
Our implementation leverages the well-established open-source monitoring stack, using Grafana for visualization and Prometheus for data provisioning.
We demonstrate its efficiency by integrating into Kaapana interfacing its I/O pipelines and externally monitored tools.
Finally, we deploy the integrated monitoring tool within the RACOON project, enabling seamless data and collaborators exploration and progress tracking in multi-centric radiology studies.

The framework and its integration into Kaapana were developed according to the requirements outlined in Section \ref{sec:2_1_requirements}.
The final solution effectively meets R1 and R2 by adopting a client-server architecture between local instances and a central server, and leveraging an established open-source monitoring stack instead of proposing a customized, single-usage data management platform.
Although demonstrated through integration with Kaapana and deployment within the RACOON consortium, the proposed monitoring architecture is platform-independent. Based on the widely adopted Grafana–Prometheus stack and standardized metric endpoints, it can be readily applied to other research infrastructures exposing monitoring metrics.
While the integration of the monitoring tool within Kaapana aligns with privacy preservation requirement R3 by ensuring that only aggregated metric data is collected rather than identifiable patient information, the tool itself does not inherently guarantee it.
Monitored tools can expose any patient-related data through their \texttt{/metrics} API endpoint, as long as it conforms to the Prometheus text-based exposition format. Consequently, the responsibility for preserving patient privacy lies in the implementation of the monitored tools themselves.
The need for configuration (R4) is well addressed through the dynamic collection of local metrics, allowing third-party systems to integrate via exposing their \texttt{/metrics} API endpoint, and the flexible visualization of monitoring data through configurable Grafana dashboards at the central server site.
However, our Kaapana integration lacks flexibility in its default metrics collection, which retrieves DICOM metadata and workflow details from OpenSearch and Apache Airflow but does not allow users to dynamically configure monitored metrics.
To minimize human intervention and reduce errors, and ensure actuality of monitored study progress, R5 requires automatic metric extraction instead of manual input into a shared table. This is achieved by running the \texttt{local-metrics-collector} workflow automatically as periodic cron jobs, ensuring metrics remain up-to-date based on the configured execution interval.
When sharing information across clinical centers, protecting both patient's data privacy and researchers' intellectual property is of essential importance.
R6 requires strict study separation, ensuring researchers can only access monitoring metrics and dashboards for their studies.
This is achieved by collecting metrics including the respective study context and deploying a separate monitoring technology stack (Prometheus Pushgateway, Prometheus, Grafana) for each study, leveraging Kaapana’s multi-installable application support for study-based separation. 
While this approach meets R6, it is resource-intensive and therefore requires future work.

A key design choice of the monitoring tool is the communication of metrics between local client sites' and central server site's Kaapana instances.
While using DICOM encapsulation via the DCMTK toolkit may seem unconventional, it ensures seamless integration with Kaapana’s default DICOM-based I/O pipeline and avoids exposing additional network ports in clinical deployments, enhancing security. However, the underlying Prometheus text-based exposition format is plain-text (\texttt{.txt}), allowing seamless adaptation to alternative communication protocols if required.

Building on the discussion of requirement fulfillment and further design choices, future releases of the proposed monitoring framework should focus on the following aspects:
Currently, the tool is built on open-source technologies and is therefore largely framework-agnostic. However, the \texttt{local-metrics-collector} and \texttt{central-metrics-extractor} workflows are tightly integrated with Kaapana and rely on its Apache Airflow workflow engine. These workflows should be restructured as standalone scripts running as cron jobs on any infrastructure, enhancing the tool's framework-agnostism.
While the tool allows flexibility in monitored system and dashboard customization, configuring monitored metrics still requires backend modifications within the monitored tool. A more user-friendly configuration system would improve usability.
Lastly, the current approach enforces strict study separation by deploying a dedicated monitoring stack per study, which is resource-intensive and may become inefficient at scale. A future release should include a more scalable solution by sharing one monitoring stack deployment across all studies, leveraging the base infrastructure’s authentication system within Prometheus and Grafana.

\section{Conclusion}\label{sec5}
Our introduced framework addresses challenges of data and collaboration exploration, and progress monitoring in multi-center radiology studies, which are often managed manually through private emails and shared tables. These practices risk becoming quickly outdated and error-prone due to manual human input, making them potentially unreliable, and unsuitable for scaling with the growing demands of large multi-center studies.
Instead, we provide a scalable, open-source, client-server implementation for up-to-date data exploration and study monitoring while ensuring patient's data privacy and protecting clinical researchers' intellectual property.
To offer accessibility and usability, we integrated our solution into the medical image processing platform Kaapana, made all implementations publicly available (\hyperlink{https://github.com/MIC-DKFZ/study-monitoring-kaapana}{https://github.com/MIC-DKFZ/study-monitoring-kaapana}), and deployed it across the Germany-wide RACOON consortium.
We encourage researchers to adopt our open-source monitoring framework on other container-based infrastructures, or deploy its Kaapana integration to streamline multi-centric studies in radiology and beyond, ultimately improving clinical outcomes and patient care.

\backmatter

\bmhead{Supplementary information}
Please refer for the Supplementary materials to the Appendix.

\bmhead{Acknowledgements}

We want to thank the whole Kaapana team of the German Cancer Research Center Heidelberg, Germany for developing the medical imaging platform Kaapana and providing it open-source.

We want to thank the engineering teams involved in the RACOON consortium for supporting in the brainstorming process of the monitored study metrics.

\section*{Declarations}

\begin{itemize}
\item Funding: This project was funded by 01KX2021 (NUM, RACOON), 01KX2121 (NUM 2.0, RACOON), 01KX2524 (NUM 3.0, RACOON).

\item Conflict of interest: Co-author Jonas Scherer is involved as a founder and CEO of Hycean GmbH. Tobias Penzkofer receives funding from Berlin Institute of Health (Advanced Clinician Scientist Grant, Platform Grant), Ministry of Education and Research (BMBF, 01KX2021 (RACOON), 01KX2121 (NUM 2.0, RACOON), 01KX2524 (NUM 3.0), 68GX21001A, 01ZZ2315D), German Research Foundation (DFG, SFB 1340/2), European Union (H2020, CHAIMELEON: 952172, DIGITAL, EUCAIM:101100633) and reports research agreements (no personal payments, outside of the submitted work) with AGO Research GmbH, Aravive, Inc., ARCAGY-GINECO, Astellas Pharma Global Development Inc., AstraZeneca AB, AstraZeneca GmbH, Clovis Oncology, EQRx International, Inc., F. Hoffmann-La Roche Ltd, GlaxoSmithKline Research \& Development Limited, Grupo Español de Investigación en Cáncer de Ovario (GEICO), ImmunoGen Inc, Incyte Corporation, Karyopharm Therapeutics, Mario Negri Gynecology Oncology Group (MaNGO) (111), Merck KGaA, Merck Sharp \& Dohme Corp., NOGGO e.V., Nordic Society of Gynaecological Oncology – Clinical Trial Unit (NSGO-CTU), Novartis Pharma GmbH, NovoCure GmbH, Sutro Biopharma, Inc., TESARO Inc., TORL Biotherapeutics, LLC, Tubulis GmbH, Universitätspoliklinik A. Gemelli, and Verastem Inc, as well as fees for a book translation (Elsevier) and speaking engagements (Bayer Healthcare). 

\item Ethics approval and consent to participate: Not applicable.

\item Consent for publication: Not applicable.

\item Data availability: Not applicable.

\item Materials availability: Not applicable.

\item Code availability: \hyperlink{https://github.com/MIC-DKFZ/study-monitoring-kaapana}{https://github.com/MIC-DKFZ/study-monitoring-kaapana}

\item Author contribution: -

\end{itemize}


\begin{appendices}






\end{appendices}


\bibliography{sn-bibliography}

\end{document}